\documentclass[conference]{IEEEtran}
\IEEEoverridecommandlockouts  
\usepackage[pdftex]{graphicx}
\usepackage{subfigure}
\usepackage{caption}
\usepackage{amsmath}
\usepackage{amssymb}
\usepackage{amsfonts}
\usepackage{algorithm}
\usepackage{algorithmic}
\usepackage{array}
\usepackage{mathtools}
\usepackage{url}
\usepackage{cite}
\begin{document}
\title{\huge Online State-Time Trajectory Planning Using Timed-ESDF in Highly Dynamic Environments}

\author{Delong Zhu, Tong Zhou, Jiahui Lin, Yuqi Fang, and Max Q.-H. Meng$^{*}$
		\thanks{$^1$The authors are with the Department of Electronic Engineering, The Chinese University of Hong Kong, Shatin, N.T., Hong Kong SAR, China. \textit{email: \{dlzhu, tzhou, jhlin, yqfang\}@ee.cuhk.edu.hk} }
		\thanks{$*$Max Q.-H. Meng (the corresponding author) is with the Department of Electronic and Electrical Engineering of the Southern University of Science and Technology in Shenzhen, China, on leave from the Department of Electronic Engineering, The Chinese University of Hong Kong, Hong Kong, and also with the Shenzhen Research Institute of the Chinese University of Hong Kong in Shenzhen, China. \textit{e-mail: max.meng@ieee.org}. This project is partially supported by the Hong Kong RGC GRF grants \#14200618 and Hong Kong ITC ITSP Tier 2 grant \#ITS/105/18FP awarded to Max Q.-H. Meng.}
	}

\maketitle
\begin{abstract}
Online state-time trajectory planning in highly dynamic environments remains an unsolved problem due to the unpredictable motions of moving obstacles and the curse of dimensionality from the state-time space. Existing state-time planners are typically implemented based on randomized sampling approaches or path searching on discretized state graph. The smoothness, path clearance, and planning efficiency of these planners are usually not satisfying. In this work, we propose a gradient-based planner over the state-time space for online trajectory generation in highly dynamic environments. To enable the gradient-based optimization, we propose a Timed-ESDT that supports distance and gradient queries with state-time keys. Based on the Timed-ESDT, we also define a smooth prior and an obstacle likelihood function that is compatible with the state-time space. The trajectory planning is then formulated to a MAP problem and solved by an efficient numerical optimizer. Moreover, to improve the optimality of the planner, we also define a state-time graph and then conduct path searching on it to find a better initialization for the optimizer. By integrating the graph searching, the planning quality is significantly improved. Experiment results on simulated and benchmark datasets show that our planner can outperform the state-of-the-art methods, demonstrating its significant advantages over the traditional ones.   
\end{abstract}

\section{Introduction}
Online motion planning for finding smooth, collision-free, and time-optimal trajectories in dynamic environments is a fundamental problem in numerous robotic applications, e.g., service robots \cite{drl-zhu, Tingguang2019Learning, chaoqunsrm}, aerial vehicles \cite{wang2020tartanair, hawkeye-zhu}, and self-driving cars \cite{urmson2008autonomous, survey2016}. However, planning in dynamic environments is a challenging task, since it has to cope with the motions of moving obstacles for avoiding collisions, which is typically a difficult and time-consuming process. Meanwhile, the constraint of time optimality on the planner results in high-dimensional searching spaces, e.g., \textit{state-time} or \textit{configuration-time} space, making the problem even more challenging. The completeness and optimality of planning with moving obstacles are practically unachievable \cite{survey2020}, and existing studies mostly focus on finding better sub-optimal solutions with the attempt to ensure the completeness under specific assumptions.  

The quality of the planned trajectory highly depends on two factors: whether the future motion of obstacles is observable, and how the motions are incorporated into planning. In practice, the future motion of obstacles, e.g., pedestrians, is typically unobservable, hence some motion prediction techniques \cite{traj-pred-2018, traj-pred-2019, forcast-2018} are developed to help improve the planning quality. Another frequently used method is \textit{velocity extrapolation} \cite{vo98}, which leverages the current velocity of obstacles to simulate their future motions within a short time interval. The simulation errors are then corrected by a fast re-planning process. The philosophy of this method is to constantly linearize the future motion of obstacles and then design collision avoidance strategies based on the linearization. In this work, we adopt the velocity extrapolation method and focus on developing an online trajectory planning framework that can improve the planning safety and fast-replanning capability.

The proposed planning framework is developed over state-time space \cite{state-time-sapce98}, which is the state space augmented with the time dimension. Compared with configuration-time space adopted by \cite{2012Trajectory, 2016Integrated}, the state-time space allows us to define high-order cost maps that can integrate the influence of high-order motions of obstacles, e.g., positions and velocities, which is the key to enable high-order collision checking. Based on such maps, we can then directly perform trajectory optimization on the state-time space considering the smoothness, safety, and time optimality simultaneously. 
There are also some studies \cite{ics-safe-planning05, dwa07, drt-17} in the literature that base the planning on state-time space in the literature, but most of them are sampling based or graph-searching based. The smoothness and clearance of the planned paths by such methods are typically not satisfying, and some extra processing are needed, e.g., connecting neighbor states with motion primitives \cite{lattice}. While the sampling based methods are effective at solving high dimensional state-time planning problems, they take a long time to converge and the quality of intermediate solutions are also unpredictable \cite{lattice}. 

In our proposed framework, we integrate a front-end graph searching algorithm for finding a near-optimal initial path and a back-end gradient-based planner for state-time trajectory optimization. In the front end, we define a state-time graph based on the geometrics of moving obstacles, and we then extend the conventional A* searching algorithm to a state-time version. Since the state-time graph is sparse, the front-end searching maintains very high efficiency. In the back end, we first propose a Timed-ESDF (Euclidean Signed Distance Field) that supports distance and gradient queries using state-time keys, and then we define a motion prior and a collision likelihood function, which are compatible with state-time space, to evaluate the smoothness and clearance of candidate trajectories, respectively. Lastly, the trajectory optimization is conducted by solving a \textit{maximum posterior} (MAP) problem. Compared with conventional methods, the proposed framework has the following advantages: 1) The formulation is based on state-time space and thus the obstacle dynamics are fully leveraged for collision avoidance; 2) The safety, smoothness, and time optimality are all considered, which significantly improves the quality of trajectories; 3) The generation of cost maps, i.e., the state-time graph and Timed-ESDF, and the optimization process are very efficient, which enables high-frequency online trajectory generation.

\section{Related Work}
Planning methods for dynamic environments can be broadly divided into three categories, i.e., reactive planners, partial planners, and global planners, according to planning horizons. 

\textbf{Reactive planners} take into account obstacles within a finite range and only computes one-step actions without long-horizon planning. Typical methods include velocity obstacles (VO) \cite{vo98, gvo09, avo11}, inevitable collision states (ICS) \cite{ics-theory04, probability-ics-12}, and artificial potential field (APF) based ones \cite{apf17}. Fiorini \textit{et al.} \cite{vo98} first propose the concept of VO, which defines obstacles in velocity space by integrating the motions of moving obstacles. Wilkie \textit{et al.} \cite{gvo09} then generalize this concept by introducing constraints of a car-like robot and propose generalized VO. Van Den Berg \textit{et al.} \cite{avo11} further extend the concept to the acceleration dimension and propose acceleration-velocity obstacles. By incorporating high-order motions of the moving obstacles, these methods perform well in dynamic environments. However, the lack of long-horizon planning makes them easily trapped in local minimums. The concept of ICS is proposed by Fraichard \textit{et al.} \cite{ics-theory04} and then is extended with a probabilistic definition \cite{probability-ics-12}. The ICS defines a set of collision states that a robot can never escape from. By avoiding ICS, the robot is guaranteed to find a collision-free action towards the goal. Malone \textit{et al.} \cite{apf17} propose the Stochastic Reachable (SR) set to model the stochastic motion of moving obstacles, and then use it to weight the potential field used for planning. The limitation of APF based methods is that the local minimum in the field may entrap the robot.

\textbf{Partial planners} only consider obstacles within a finite time horizon or sensing range, and the planned trajectory may not reach the goal. Partial motion planning (PMP) is first proposed by Petti \textit{et al.} \cite{ics-safe-planning05}, based on such an idea that the dynamic environment imposes real-time constraints on motion planning and it is impossible to compute a complete trajectory within the time available. Therefore, the authors focus on computing the best partial trajectory with the guarantee that the end state of the trajectory is ICS-free. Benenson \textit{et al.} \cite{ics-safe-planning06} then apply the PMP to autonomous navigation task of urban vehicles, demonstrating the safety of PMP in tackling partially observable environments. The idea of PMP is widely adopted by sampling-based methods. 
The dynamic window based method proposed by Seder \textit{et al.} \cite{dwa07} is another type of partial planner. The authors first leverage a focused D* to search a reference path without considering obstacle motions, and then a dynamic window algorithm is used to generate admissible local trajectories around the reference path. Such a two-layer planning schema is also adopted by Chiang \textit{et al.} \cite{drt-17}, in which the RRT and forward SR sets are utilized to find the reference path and avoid moving obstacles, respectively. Compared with reactive planners, the partial planner integrates more long-term information and is with a better capability of avoiding local minimums. The partial planner is also a compromise between planning efficiency and optimality.

\textbf{Global planners} take into account all obstacles in the environment and plan trajectories that connect the start and goal. Fraichard \textit{et al.} \cite{state-time-sapce98} for the first time propose the concept of \textit{state-time} space, which is then taken as a tool to formulate the planning problem in a dynamic workspace. Based on the formulation, a state-time A* is proposed to search near-time-optimal solutions in the state-time space. Sintov \textit{et al.} \cite{timed-rrt} reformulate the conventional RRT algorithm on the state-time space and propose a Time-Based RRT algorithm, in which a time dimension is added to the tree so that each node denotes a specific state at a specific time. The global solution is then obtained by explosive node expansion. Although the optimality of these methods is well ensured, their inefficiency makes them only suitable for small-scale problems.  
In a recent study, Cao \textit{et al.} \cite{dynamic-channel} propose the concept of \textit{dynamic channel} (DC) based on Delaunay triangulation, in which each edge is defined as a navigational gate, and the gate sequence that connects the start and goal is then defined as a dynamic channel. Accordingly, the planning task is converted to finding the shortest and collision-free channel on the triangle graph. Different from \cite{state-time-sapce98} and \cite{timed-rrt} that directly discrete the state-time space, the DC method adopts a more sparse representation of the space and thus shows a much higher planning efficiency.

The proposed method in this work falls into the category of global planners, and the contributions are as follows:
\begin{itemize}
	\item We propose an online trajectory planning framework on the state-time space, which can generate smooth, collision-free, and near-time-optimal trajectories with a frequency of 30-50Hz.	
	\item We propose a state-time graph for fast searching of the initial path and a Timed-ESDF that integrates obstacle motions for online trajectory optimization.  
	\item We formulate the trajectory optimization into a MAP problem by leveraging a state-time compatible smooth prior and obstacle likelihood function. This is also the first ESDF-based state-time planner used for highly dynamic environments.  
\end{itemize}

\section{Problem Definition}
In this work, we address the motion planning problem for a holonomic mobile robot $\mathcal{A}$ in a planar workspace $\mathcal{W}$ cluttered up with moving obstacles $\mathcal{B}$. The goal is to plan an optimal trajectory $\boldsymbol{\xi}$ that achieves the planning objective $\mathcal{F}$ in the state-time space $\mathcal{ST}$. The problem is defined by a tuple $(\mathcal{W}, \mathcal{ST}, \mathcal{A}, \mathcal{B}, \mathcal{F})$, where

\begin{itemize}
	\item $\mathcal{W} \in \mathbb{R}^2$ denotes the workspace that contains the robot and moving obstacles.
	\item $\mathcal{ST}$ is the state space $\mathcal{S}$ of the robot augmented by a time dimension $\mathcal{T} $. 

	\item $\mathcal{A} \subset \mathcal{W}$ is a closed region occupied by the robot in the workspace. $\mathcal{A}(\mathbf{s},t)$ defines a mapping from the state-time space of the robot to the workspace. $\mathcal{A}^{-1}$ defines an inverse mapping.     
	
 	\item $\mathcal{B} \subset \mathcal{W}$ denotes a set of obstacles $\mathcal{B}_i, i\in \{1, \cdots, M\}$ moving in the workspace (static obstacles are not considered). Collision happens if the robot $\mathcal{A}$ intersects $\mathcal{B}_i$ at time $t$, i.e., $\mathcal{A}(\mathbf{s}, t) \cap \mathcal{B}_{i}(t) \neq \emptyset$. The inadmissible region of a state-time space is as follows, 
 	\begin{equation}
 	\small
 	\mathcal{T B}=\left\{(\mathbf{s}, t) \,|\, \exists i \in\{1, \ldots, M\}, \mathcal{A}(\mathbf{s}, t) \cap \mathcal{B}_{i}(t) \neq \emptyset\right\}.
 	\end{equation}
	\item $\mathcal{J}: \Xi \rightarrow \mathbb{R}$ denotes the objective function, which maps the trajectory $\boldsymbol{\xi}$ from trajectory space $\Xi \subset \mathcal{ST}$ to a real number. 
\end{itemize}

Based on the above definition, we formulate state-time planning in dynamic environments into an optimization problem: 
\begin{equation}
\begin{aligned}
\boldsymbol{\xi} = \arg\min_{\boldsymbol{\xi} \in \Xi}& \; \mathcal{J}_{prior}[\boldsymbol{\xi}] + \mathcal{J}_{obs}[\boldsymbol{\xi}] + \mathcal{J}_{time}[\boldsymbol{\xi}]\\
\mbox{s.t.}\quad
&\boldsymbol{\xi}(t_0) = \mathbf{s}_0,  \boldsymbol{\xi}(t_f) = \mathbf{s}_f, \\
& \boldsymbol{\xi}(t) \in \mathcal{ST} \, \backslash \, \mathcal{TB}.\\
\end{aligned}
\label{problem}
\end{equation}
$\mathcal{J}_{prior}, \mathcal{J}_{obs}$, and $\mathcal{J}_{time}$ measure the smoothness, safety, and time cost of a trajectory, respectively. $\mathbf{s}_0$ and $\mathbf{s}_f$ are the start and goal states, respectively. $\mathcal{AST}\small{=}\,\mathcal{ST} \backslash \mathcal{TB}$ defines an \textit{admissible state-time space}, in which the feasible trajectories are embedded. 
It is worth noting that $\mathcal{J}_{prior}$ and $\mathcal{J}_{obs}$ depend on a fixed time assignment, which contradicts the goal of optimizing $\mathcal{J}_{time}$. To address this problem, we adopt a two-step optimization procedure. A state-time A* is first leveraged to find a near-time-optimal path, and then taking this path as initialization, a trajectory optimization method is applied to generate high-quality trajectories.

\section{Front-end Path Searching}
The front end performs collision-free and near-time-optimal path searching on a timed triangle graph by assuming piece-wise constant velocity for the robot, i.e., $\mathcal{J}_{prior}$ is relaxed. We start by presenting the graph building process and then detail the state-time A* algorithm. 
\subsection{State-Time Graph Building}
\begin{figure}[t]
	\centering
	\subfigure[The timed triangle gaph at $t$ and $t+d_t$.]{
		\label{ttga}
		\includegraphics[width=0.22\textwidth]{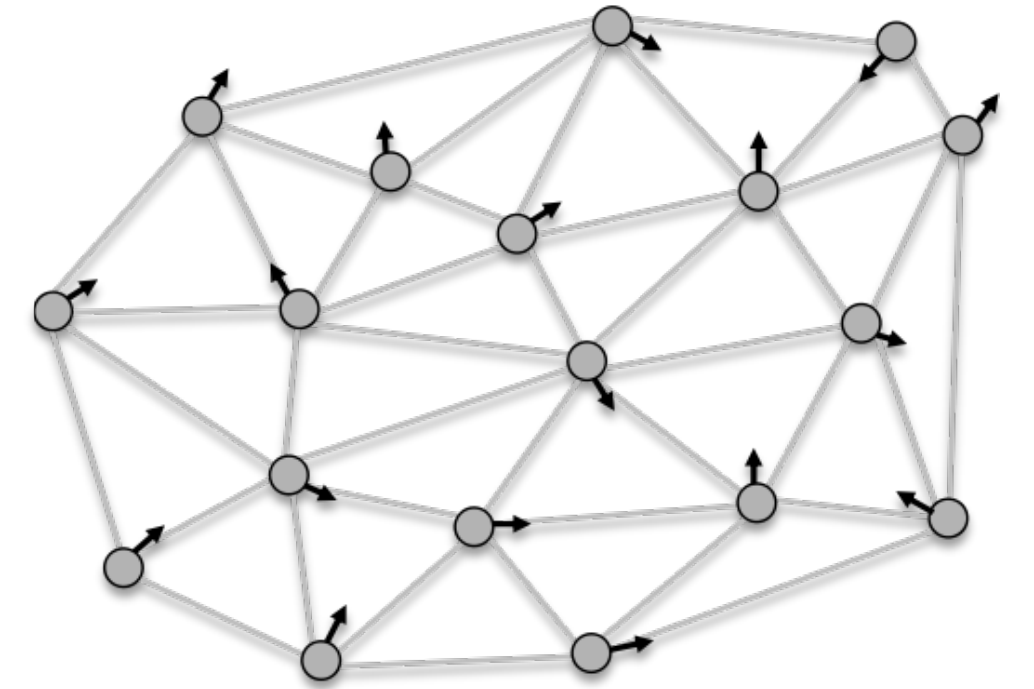}
		\includegraphics[width=0.22\textwidth]{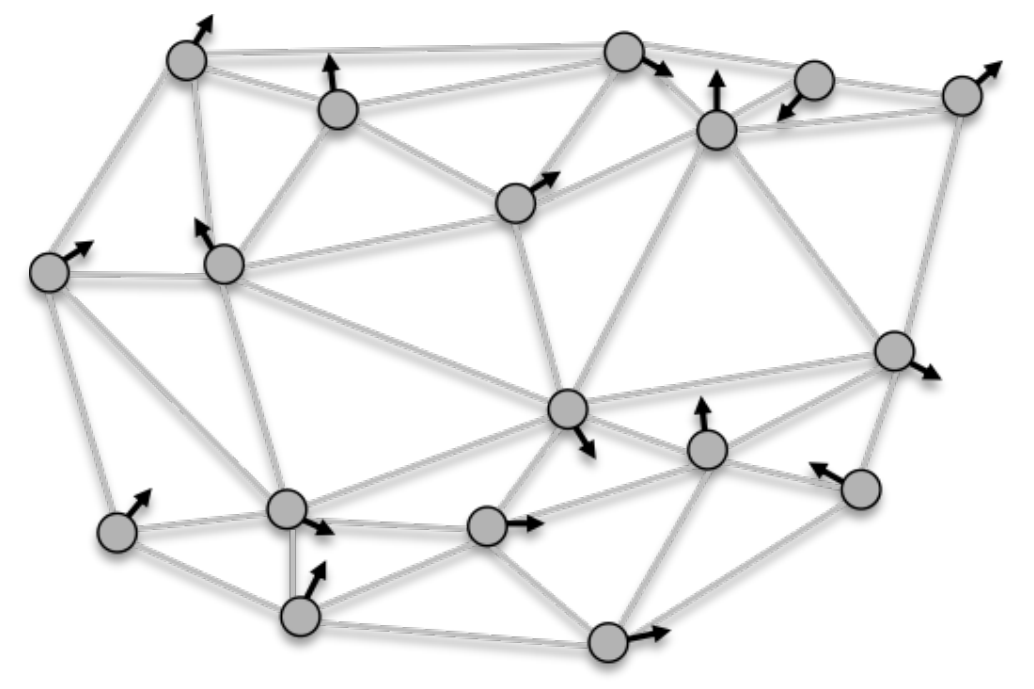}
	} 
	
	\subfigure[The state-time dual gaph at $t$ and $t+d_t$.]{
		\label{ttgd}
		\includegraphics[width=0.22\textwidth]{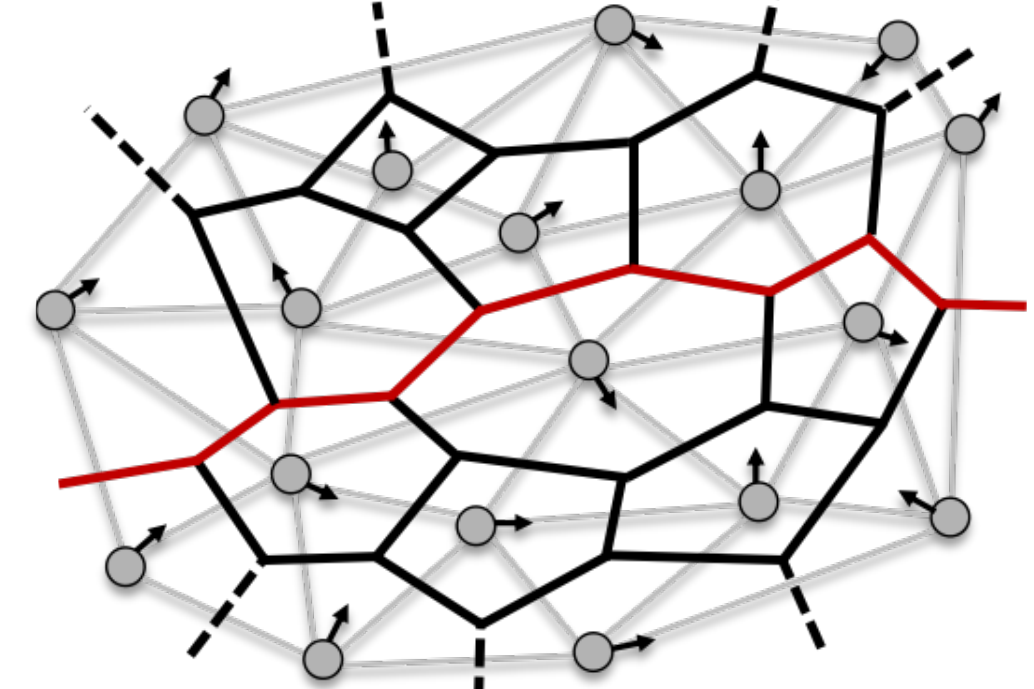}
		\includegraphics[width=0.22\textwidth]{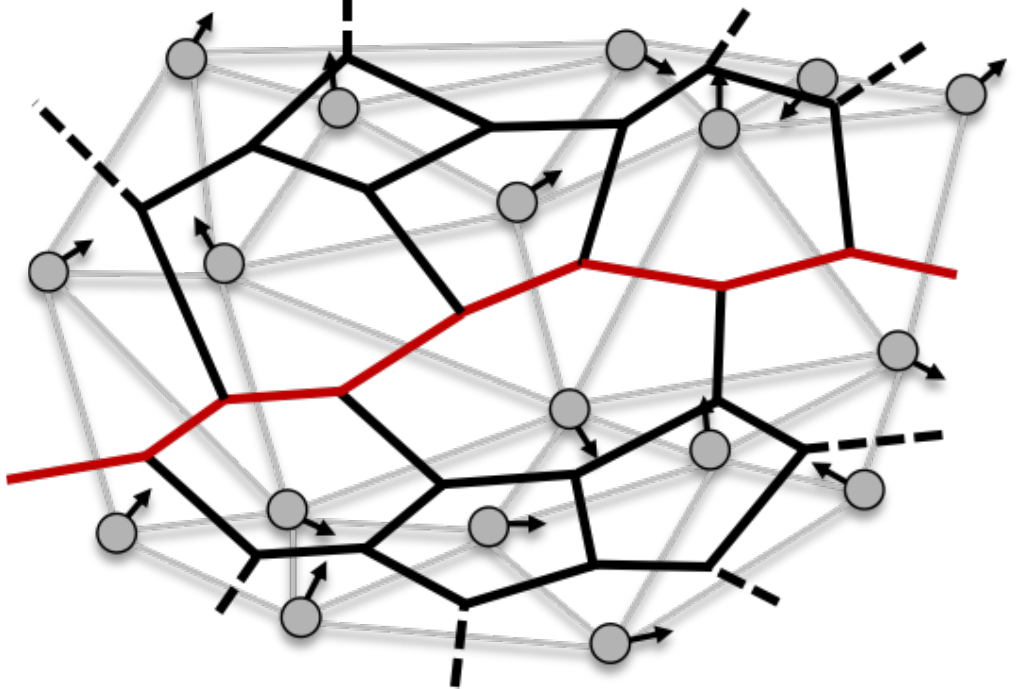}
	}
	\caption{Demonstrations of the timed triangle graph and its dual graph. The red segments indicate a feasible path across the moving obstacles.}
	\label{fig:ttg}
	\vspace{-2mm}
\end{figure}
\textbf{Graph definition}: The \textit{timed triangle graph} $\mathcal{G}= \,<$$\mathcal{V}(t)$, $\mathcal{E}(t)$, $\mathcal{F}(t)$$>$, shown in Fig. \ref{ttga}, is a topological representation of the moving obstacles in workspace, where $\mathcal{V}(t),\mathcal{E}(t), \mathcal{F}(t)$ are vertices, edges, and faces of the graph at time $t$. Each vertex $\mathcal{V}_i$ encodes a future trajectory of $\mathcal{B}_i$, which is obtained by motion forecasting of the obstacle. Under the assumption of constant velocity, $\mathcal{V}_i$ encodes a straight-line trajectory of $\mathcal{B}_i$. The edge $\mathcal{E}(t)$ and face $\mathcal{F}(t)$, which depend on $\mathcal{V}(t)$, are generated by the {Delaunay triangulation} at time $t$. The edge $\mathcal{E}(t)$ splits the workspace $\mathcal{W}$ into multiple triangle faces and each face $\mathcal{F}_i(t)$ defines a geometric group of collision-free regions in $\mathcal{W}$. Accordingly, $\mathcal{F}_i$ induces an \textit{admissible triangle} $\mathcal{AF}_i = \mathcal{AST} \cap \mathcal{A}^{-1}(\mathcal{F}_i)$. Path searching on $\mathcal{G}$ is essentially to find a sequence of admissible triangles that can constitute a safe channel for the robot to navigate through. However, $\mathcal{G}$ is defined in a continuous space, which makes it not suitable for path searching. Therefore, we need to build a discretized representation for $\mathcal{G}$. 

\textbf{Dual graph}: We first discrete the time dimension of $\mathcal{G}$ into $T$ slices with a resolution $d_t$, as shown in Fig. \ref{ttga}. Each slice is a Delaunay triangulation with a time index $t \in \{1, \cdots, T\}$, denoted by $\mathcal{G}[t]$, which indicates an instantaneous topology between moving obstacles. We then simplify each triangle into a node $\mathcal{AN}_i[t] \in \mathcal{AF}(t)$ and build feasible connections $\mathcal{AE}[t] \in \mathcal{AF}(t)$ for adjacent nodes. As a result, the dual graph is constructed, $\mathcal{G}^{\dagger}[t]=<\mathcal{AN}[t], \mathcal{AE}[t]>, t \in \{1, \cdots, T\}$. As is demonstrated, two slices of the dual graph are visualized in Fig. \ref{ttgd}. It is worth noting that the dual graph is not unique, since the placements for each node are different and the node connections are also diverse. The graph building in this work is to generate an optimal dual graph under specific criteria.


\textbf{State-time graph building}: Suppose that the current state-time of the robot is $(\mathbf{p}_{r}, \mathbf{v}_{r}, t) \in \mathcal{AF}(t)$, which is a node that represents the current triangle, the problem of placing a node with admissible connections for each of the neighbor triangles is then formulated as follows: 
\begin{equation}
  \begin{aligned}
	\max_{\vec{\mathbf{v}}, t} \quad & \vec{\mathbf{p}}_{rg}^T \vec{\mathbf{v}} + \mathbf{v} t\\
	\mbox{s.t.}\quad
	& 0< t < t_m,  \\
	&\mathbf{0} < \mathbf{p}_{r} + \mathbf{v}t< \mathbf{p}_{m}, \\
	&|\mathbf{v}|  \leq \mathbf{v}_{m}, \; \mathbf{p}_{sg}^T \mathbf{v} > 0, \\
	&\left\|\mathbf{p}_{ri} + (\mathbf{v}-\mathbf{v}_i)t \right\|_2 > C_s,\forall i \in \mathcal{V}(t),
  \end{aligned}
  \label{placenode}
\end{equation}
where $\vec{\mathbf{x}}$ denotes the unit vector of $\mathbf{x}$, $(\mathbf{p}_{m}, \mathbf{v}_{m}, t_m)$ provides the state-time boundary of a triangle, $C_s$ is the radius of obstacles, and $\mathbf{p}_{rg}, \mathbf{p}_{sg}, \mathbf{p}_{ri}$ are the robot-goal vector, start-goal vector, and robot-obstacle vector, respectively. The objective of Eq. \eqref{placenode} is to encourage a long-distance movement of the robot towards the goal, and the endpoint of the movement is taken as the target node placement $(\mathbf{p}_{r}\small{+}\mathbf{v}^* t, \mathbf{v}^*, t^*)$. The safety, or admissibility, of the connection between the current and the target node is ensured by the last constraint in Eq. \eqref{placenode}. 

Solving Eq. \eqref{placenode} with numerical methods is computationally prohibitive, since $\vec{\mathbf{v}}$ and $t$ are coupled. In this work, we choose to discrete $\vec{\mathbf{v}}$ and $t$, and conduct forward simulation to test the constraints and optimize the objective function. Suppose there are $P$ candidate velocities and $Q$ candidate time points after discretization, and $M$ moving obstacles, the time complexity of solving Eq. \eqref{placenode} is $\mathcal{O}(PQM)$. Due to the sparsity of $\mathcal{G}^{\dagger}$, the graph can be constructed very efficiently.  
   
\subsection{State-time A*}
\begin{algorithm}[t]
	\caption{State-time A*(\textit{start}, \textit{goal}, $T$, $\mathcal{G}$)}
	\label{astar}
		\begin{algorithmic}[1]
			\STATE	\textsc{OPEN} $\gets \emptyset$, \\
			\STATE \textsc{CLOSED} $\gets \emptyset$,  \\ 
			\STATE	g(\textit{start}) $\gets$ 0 \\
			\STATE	\textsc{OPEN.Insert}(\textit{start}, $t_0$) \\	
			\WHILE{\textsc{OPEN} $\not =\emptyset$ }{
				\STATE ($\mathbf{s}, t$) $\gets$ OPEN.\textsc{Pop()} \\
				\STATE ($\mathbf{s}_g,\, t_g$) $\gets$ \textsc{GoalReached}($\mathbf{s}, t$) \\
				\STATE ($\mathbf{s}_n, t_n$) $\gets$ \textsc{GoalNearest}($\mathbf{s}, t$) \\
				\IF{$\mathbf{s}_g = $ goal}
					\RETURN \textsc{BackTrack}($\mathbf{s}_g,\, t_g$) 
				\ELSIF{$t_g \geq T$ }
					\RETURN \textsc{BackTrack}($\mathbf{s}_g,\, t_g$) 
				\ENDIF
				\STATE /* \textit{get neighbor triangles in the current graph} */  \\
				\FOR{$\Delta_t \in$ \textsc{NeighborTriangles}($\mathbf{s}, \mathcal{G}[t]$)}
					\STATE /* \textit{place a node for $\Delta_t$ by solving Eq. \eqref{placenode}} */  \\
					\STATE ($\mathbf{s}', t'$) $\gets$ \textsc{PlaceNode}($\mathbf{s}, \Delta_t, \mathcal{G}$)\\
					\IF{g($\mathbf{s}'$) $>$  g($\mathbf{s}'$) + \textsc{NaviCost}($\mathbf{s},\mathbf{s}'$)}
						\STATE g($\mathbf{s}'$) $\gets$ g($\mathbf{s}'$) + \textsc{NaviCost}($\mathbf{s},\mathbf{s}'$) \\
						\STATE h($\mathbf{s}'$) $\gets$ \textsc{EuclideanDist}($\mathbf{s}'$, \textit{goal}) \\
						\STATE \textsc{OPEN.Insert}(\textit{$\mathbf{s}', t'$}) 
					\ENDIF	   
				\ENDFOR
				\STATE \textsc{CLOSED.Insert}($\mathbf{s}, t$) \\
			}
			\ENDWHILE
			\RETURN \textsc{BackTrack}($\mathbf{s}_n,\, t_n$)\\			
		\end{algorithmic}	
\end{algorithm}
Based on the definition of $\mathcal{G}^{\dagger}$, a state-time A* algorithm is leveraged to search the near-time-optimal path, shown in Alg. \ref{astar}. The differences with a standard A* mainly lie in three aspects. Firstly, a partial solution is allowed. Once the time budget $T$ is used up or the searching process fails, the current optimal path is returned (lines 7-13). As mentioned in graph definition, $\mathcal{V}_i$ encodes a predicted trajectory of $\mathcal{B}_i$, which means $\mathcal{G}^{\dagger}$ cannot provide accurate information for long-term planning. Therefore, a partial solution is acceptable. There is a high possibility that a complete path is found during the execution of the partial one. 
Secondly, the neighbor retrieval, i.e., \textsc{Succ}, is based on $\mathcal{G}[t]$, and the node is placed on the corresponding triangle in $\mathcal{G}[t']$ (lines 15-17). In this way, the connections across different time slices are built up.
Thirdly, the node placement is performed along with the progress of path searching, which helps avoid a lot of unnecessary node connections. Finally, a \textsc{NaviCost} is defined to incorporate the instantaneous collision potentials,
\begin{equation}
	J_{\rm navi}=
	\begin{cases}
	d(\mathbf{s},\mathbf{s}'),& d(\mathbf{s},\mathbf{s}') > d_0, \\
	d(\mathbf{s},\mathbf{s}')*(2 - {d(\mathbf{s},\mathbf{s}') }/{d_0}),& d(\mathbf{s},\mathbf{s}') \leq d_0,
	\end{cases}
\end{equation}
where $d_0=2C_s$, and $d(\mathbf{s},\mathbf{s}')$ is the Euclidean distance between the two nodes. Here, $J_{\rm navi}(\mathbf{s},\mathbf{s}') \geq 	d(\mathbf{s},\mathbf{s}')$ always holds, which ensures the heuristic is admissible.

The time complexity of the algorithm is $\mathcal{O}(PQM^3\log{M})$, including the complexity of A* searching $\mathcal{O}(M\log{M})$, \textsc{Succ} complexity $\mathcal{O}(M)$, and \textsc{PlaceNode} complexity $\mathcal{O}(PQM)$. The searching on $\mathcal{G}^{\dagger}$ is complete and optimal if and only if the assumption of constant velocity for moving obstacles stands, which however, is hard to fulfill in practice. Hence, only a partial and near-optimal solution can be found. This is also a limitation of motion planning in dynamic environments.

\section{Back-end Trajectory Optimization}
The design of back-end optimizer aims to generate a smooth and safe trajectory based on the initial path planned by the front end. To achieve this goal, we first define a space of smooth trajectories based on GP, and then leverage a likelihood function to evaluate the safety of each trajectory in the space. Trajectory optimization is finished by solving a MAP problem.   

\subsection{Prior Distribution of Smooth Trajectories}
To parameterize smooth trajectories, a GP prior is defined over the trajectory space $\Xi$, and each trajectory $\boldsymbol{\xi} \in\Xi$ is a sample from this prior distribution, i.e.,   
$\boldsymbol{\xi} \sim \mathcal{G} \mathcal{P}(\boldsymbol{\mu}, \boldsymbol{\mathcal{K}})$, where $\boldsymbol{\mu}$ is the mean trajectory and $\boldsymbol{\mathcal{K}}$ is the covariance function that characterizes the shape of trajectories in the space.
Similar to previous work on GP-based motion planning \cite{gp-prior-dong, gp-prior-tim}, we consider a GP generated by a \textit{linear time varying stochastic differential equation} (LTV-SDE),
\begin{equation}
\label{system}
\dot{\boldsymbol{\xi}}(t)=\mathbf{A}(t)\boldsymbol{\xi}(t)+\mathbf{u}(t)+\mathbf{F}(t) \mathbf{w}(t),
\end{equation}
where $\boldsymbol{\xi}(t)$ is a state-time on the trajectory, $\mathbf{u}(t)$ is an exogenous input, which is assumed to be zero in this paper, $\mathbf{A}(t), \mathbf{F}(t)$ are time-varying matrices, and $\mathbf{w}(t)$ is white process noise,
\begin{equation}
\mathbf{w}(t) \sim \mathcal{G} \mathcal{P}\left(\mathbf{0}, \mathbf{Q}_{C} \delta\left(t-t^{\prime}\right)\right),
\end{equation}
where $\mathbf{Q}_{C}$ is the power-spectral density matrix and $\delta\left(t-t^{\prime}\right)$ is the Dirac delta function. A general solution to the LTV-SDE is given by 
\begin{equation}
\small
\boldsymbol{\xi}(t)=\mathbf{\Phi}\left(t, t_{0}\right) \boldsymbol{\xi}\left(t_{0}\right)+\int_{t_{0}}^{t} \mathbf{\Phi}(t, x)(\mathbf{u}(x)+\mathbf{F}(x) \mathbf{w}(x)) dx,
\label{solution}
\end{equation}
where $\mathbf{\Phi}\left(t, x\right) = \exp{\int_{x}^{t}\mathbf{A}(x)dx}$ is the transition matrix. Eq. \eqref{solution} essentially defines a stochastic process that follows the system model given by Eq. \eqref{system}.

The mean and covariance (or kernel) function of $\boldsymbol{\xi}(t)$ is as follows \cite{gp-prior-tim},
	\begin{equation}
\label{mean}
\boldsymbol{\mu}(t)=E[\mathbf{x}(t)]=\mathbf{\Phi}\left(t, t_{0}\right) \boldsymbol{\mu}_{0}+\int_{t_{0}}^{t} \mathbf{\Phi}(t, s) \mathbf{u}(s)ds,
\end{equation}
\begin{equation}
\begin{array}{l}
\boldsymbol{\mathcal{K}}\left(t, t^{\prime}\right)=\mathbf{\Phi}\left(t, t_{0}\right) \boldsymbol{\mathcal{K}}_{0} \mathbf{\Phi}\left(t^{\prime}, t_{0}\right)^{\top} \\
\quad+\int_{t_{0}}^{\min \left(t, t^{\prime}\right)} \mathbf{\Phi}(t, s) \mathbf{F}(s) \mathbf{Q}_{C} \mathbf{F}(s)^{\top} \mathbf{\Phi}\left(t^{\prime}, s\right)^{\top}ds,
\end{array}
\label{cov}
\end{equation}
where $\boldsymbol{\mu}_{0}$ is the initial mean value of the first state on the trajectory and $\boldsymbol{\mathcal{K}}_{0}$ is the covariance. The GP prior distribution is then defined with its mean $\boldsymbol{\mu}$ and covariance $\boldsymbol{\mathcal{K}}$,
\begin{equation}
\label{densefunction}
P(\boldsymbol{\xi}) \propto \exp \left\{-\frac{1}{2}\|\boldsymbol{\xi}-\boldsymbol{\mu}\|_{\boldsymbol{\mathcal{K}}}^{2}\right\}.
\end{equation}

Suppose a near-optimal time assignment, $t_{0}<t_{1}<\cdots < t_{i}<\cdots <t_{F}, t_{i}=t_0+d_t*i$, is found by the front end, then $\boldsymbol{\xi}$ is  discretized into $F+1$ waypoints accordingly. In general, each waypoint $\boldsymbol{\xi}(t_i)$ is correlated with others, and the covariance matrix $\boldsymbol{\mathcal{K}} \in \mathbb{R}^{F\times F}$ is dense. However, it is proved that LTV-SDE is with Markov property \cite{gp-prior-tim}, and $\boldsymbol{\mathcal{K}}$ has a very sparse structure. Eq. \eqref{densefunction} thus can be simplified to a more efficient formula,
\begin{equation}
\begin{aligned}
&P(\boldsymbol{\xi}) \propto \exp \left\{-\frac{1}{2}  \sum_{i}  \boldsymbol{e}_{i}^{T} \mathbf{Q}_{i}^{-1} \boldsymbol{e}_{i}\right\}, \\
\boldsymbol{e}_{i}&= \;\mathbf{u}(t_i)-\boldsymbol{\xi}(t_i)+\boldsymbol{\Phi}\left(t_{i}, t_{i-1}\right) \boldsymbol{\xi}(t_{i-1}), \\
\mathbf{Q}_{i}&=\int_{t_{i-1}}^{t_{i}} \boldsymbol{\Phi}\left(t_{i}, x\right) \mathbf{F}(x) \mathbf{Q}_{c} \mathbf{F}(x)^{\top} \boldsymbol{\Phi}\left(t_{i}, x\right)^{\top} d x,
\end{aligned}
\label{markov}
\end{equation}
which actually measures the difference between the actual state $\boldsymbol{\xi}(t_i)$ and the expected state $\mathbf{u}(t_i)+\boldsymbol{\Phi}\left(t_{i}, t_{i-1}\right) \boldsymbol{\xi}(t_{i-1})$ by the system w.r.t $\mathbf{Q}_{i}$. The smaller the difference is, the less system changes or control efforts are needed, and thus the smoother the trajectory is. 

Compared with conventional smoothness techniques \cite{chomp09, itomp12, chomp-poly, chomp-bspline}, a significant advantage of the presented method is that the smoothness is directly measured in the state-time space, i.e., each point in $\boldsymbol{\xi}$ is a state-time, which allows us to optimize the position and velocity simultaneously. As is studied in \cite{survey2020}, the optimization of robot velocity is crucial to improve the success rate of planning with moving obstacles.

\subsection{Timed-ESDF}
\begin{figure}[t]
	\centering		
	\subfigure[The traditional ESDF with position-based queries.]{
		\label{tsdf}
		\includegraphics[width=0.33\textwidth]{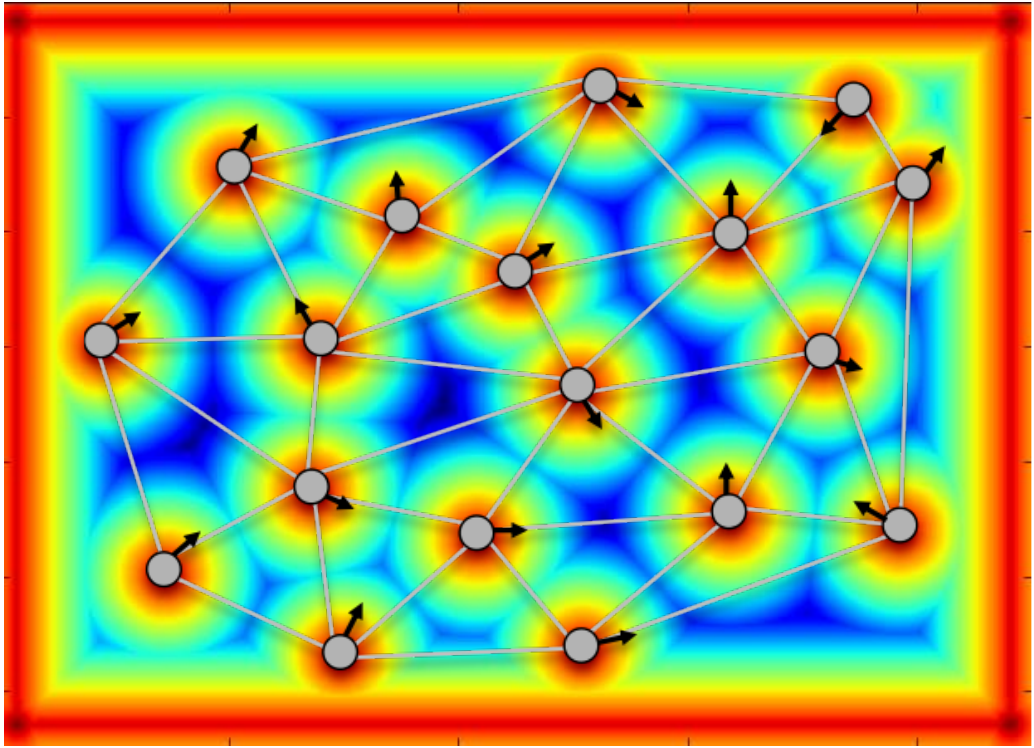}
	} 
	
	\subfigure[A slice of the Timed-ESDF with state-time queries. ]{
		\label{tesdf}
		\includegraphics[width=0.33\textwidth]{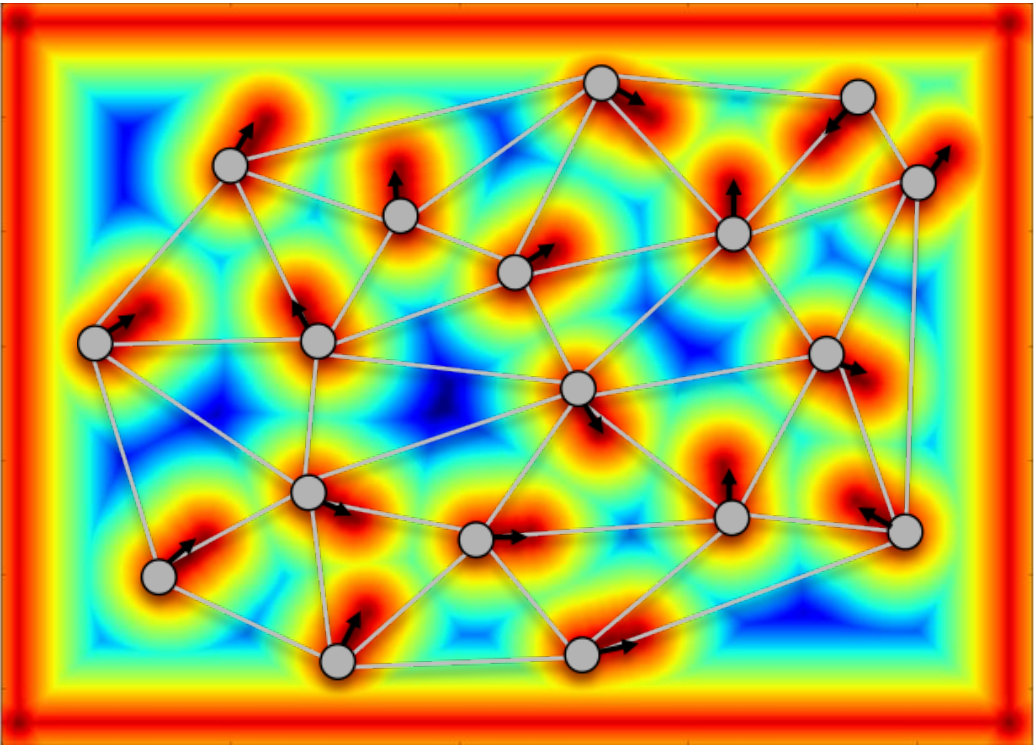}
	}
	\caption{Demonstration of traditional ESDF and the Timed-ESDF.}
	\label{fig:esdf}
	\vspace{-4mm}
\end{figure}
ESDF is a frequently used cost map, shown in Fig. \ref{tsdf}, that contains the distance and gradient information against obstacles. Timed-ESDF, shown in Fig. \ref{tesdf}, is an extended ESDF augmented with the time dimension. Based on the distance information in Timed-ESDF, we can define a likelihood function to evaluate the clearance of a trajectory,
\begin{equation}
\label{likelihood}
P(\boldsymbol{\xi}(t_i) \in \mathcal{AST} | \boldsymbol{\xi}(t_i)) \propto \exp \left\{-\frac{1}{2}\left\|\mathbf{h}\left(\boldsymbol{\xi}(t_i)\right)\right\|_{\boldsymbol{\Sigma}_{{i}}}^{2}\right\},
\end{equation}
\begin{equation}{h}(\boldsymbol{\xi}(t))=\left\{\begin{array}{cl}
-d(\mathbf{s}, t)+\epsilon & \text { if } d(\mathbf{s}, t) \leqslant \epsilon ,\\
0 & \text { if } d(\mathbf{s}, t)>\epsilon,
\end{array}\right.\end{equation}
where $d(\mathbf{s}, t)$ is the signed distance of $\mathbf{s}$ in ESDF on slice $t$, and $\epsilon$ is the safety margin. By differentiating Eq. \eqref{likelihood} w.r.t. waypoint $\mathbf{s}$ on the target slice $t$, we can get the gradient that tends to push the trajectory away from obstacles.

In addition to the time dimension, another major extension of Timed-ESDF lies in the integration of velocity obstacles \cite{vo98, gvo09}, which is the key to enable state-time planning. The conventional ESDF only supports position-based queries and is not compatible with planning in the velocity dimension. Moreover, the conventional ESDF cannot provide safety guarantee between adjacent waypoints, i.e., the interval $(\boldsymbol{\xi}(t_i), \boldsymbol{\xi}(t_{i+1}))$, in dynamic environments, since the safety optimization is only performed on the discretized waypoints rather than the entire trajectory. Accordingly, the second improvement of Timed-ESDF is to address this problem. The method is as follows.

We first give the definition of \textit{velocity obstacle} \cite{vo98, gvo09},
\begin{equation}
\mathcal{V O}_{r | i}=\left\{\mathbf{v} | \exists t>0, \, \mathbf{p}_{r}+t(\mathbf{v}-\mathbf{v}_{i}) \in \mathcal{B}_k\right\},
\label{vo}
\end{equation}
which can be equivalently written as the following formula,
\begin{equation}
\small
\mathcal{V O}_{r | i}=\left\{\mathbf{v} | \exists t>0, \, ||(\mathbf{p}_{r}+\mathbf{v}t)-(\mathbf{p}_{k}+\mathbf{v}_{k}t)||_2 \leq C_s\right\},
\label{vot}
\end{equation}
where $\mathbf{p}_{r}$ and $\mathbf{v}$ are the position and velocity of the robot, $C_s$ and $\mathbf{p}_{k}$ are the radius and center of obstacle $\mathcal{B}_k$, respectively. The set $\{\mathbf{v} | \exists t>0, \mathbf{p}_{k}+\mathbf{v}_{k}t\}$ and $C_s$ then defines the center and radius of a trajectory of $\mathcal{B}_k$, denoted by $\boldsymbol{\xi}({\mathcal{B}_k})$. Eq. \eqref{vot} indicates that there should be no overlap between the robot and $\boldsymbol{\xi}({\mathcal{B}_k})$ at any time step $t$, in order to avoid collisions.

We then give a more strict safety criterion and extend the concept of velocity obstacle as follows, 
\begin{equation}
\small
\mathcal{V O}_{r | i}(t_i, t_{i+1})=\left\{\mathbf{v} | \exists t\in [t_i, t_{i+1}), \, \mathbf{p}_{r}+\mathbf{v}t\in {L}({\mathcal{B}_k}) \right\},
\label{vt}
\end{equation}
where ${L}({\mathcal{B}_k})$ is a line segment in the workspace corresponding to $\boldsymbol{\xi}({\mathcal{B}_k})$, as shown in Fig. \ref{tesdf}. The meaning of Eq. \eqref{vt} is that the waypoint $\boldsymbol{\xi}(t_i)\small{=}(\mathbf{p}_{r}, \mathbf{v}, t)$ should never overlap with ${L}({\mathcal{B}_k})$.
Accordingly, by leveraging $\mathbf{p}_{r}+\mathbf{v}t$, we can directly query the distance in Timed-ESDF, and the gradient w.r.t $\mathbf{p}_{r}, \mathbf{v}$ can also be derived. In this way, the state-time based query is enabled, and the safety along the trajectory is ensured.

\subsection{MAP Trajectory}
Based on the GP prior and likelihood function, the back-end trajectory optimization is then completed by solving a MAP problem \cite{gp-prior-dong},
\begin{equation}
\small
\begin{aligned}
\boldsymbol{\xi}^{*}& =\underset{{\boldsymbol{\xi}}}{\operatorname{argmax}}\left\{P(\boldsymbol{\boldsymbol{\xi}}) \prod_{i} P(\boldsymbol{\xi}(t_i) \in \mathcal{AST} | \boldsymbol{\xi}(t_i)) \right\} \\
&=\underset{{\boldsymbol{\xi}}}{\operatorname{argmin}}\left\{-\log \left(P(\boldsymbol{\boldsymbol{\xi}}) \prod_{i} P\left(\boldsymbol{\xi}(t_i) \in \mathcal{AST} | \boldsymbol{\xi}(t_i)\right)\right)\right\} \\
&=\underset{\boldsymbol{\xi}}{\operatorname{argmin}}\left\{\frac{1}{2}\|\boldsymbol{\xi}-\boldsymbol{\mu}\|_{\boldsymbol{\kappa}}^{2}+\frac{1}{2}\|\boldsymbol{h}(\boldsymbol{\xi})\|_{\boldsymbol{\Sigma}}^{2}\right\}.
\end{aligned}
\label{opt}
\end{equation}
There are many numerical optimizers for solving Eq. \eqref{opt}, e.g., the Gauss-Newton or Levenberg-Marquardt optimizer. However, the accuracy and success rate of numerical methods highly depend on initial values. This is also the reason why a front-end path searching is needed.

\section{Experiments}
In this section, we first conduct control experiments on a simulation environment under different parameter settings to study performance changes of the proposed method. We then perform an overall evaluation on three benchmark datasets. During the experiments, three baseline methods are adopted: the passive \textit{wait-and-go} (WG) strategy, the classical \textit{velocity obstacle} (VO) method \cite{gvo09}, and the state-of-the-art \textit{dynamic channel} (DC) method \cite{dynamic-channel}. Our method is denoted by \textit{state-time planner} (ST). The experiment platform is a laptop with i7-7700HQ 2.8GHZ CPU and 16GB memory.

\subsection{Experiments on Simulation Environment}

\begin{figure}[t]
	\centering
	\subfigure[Multi-robot simulation.]{
		\label{sim}
		\includegraphics[width=0.22\textwidth]{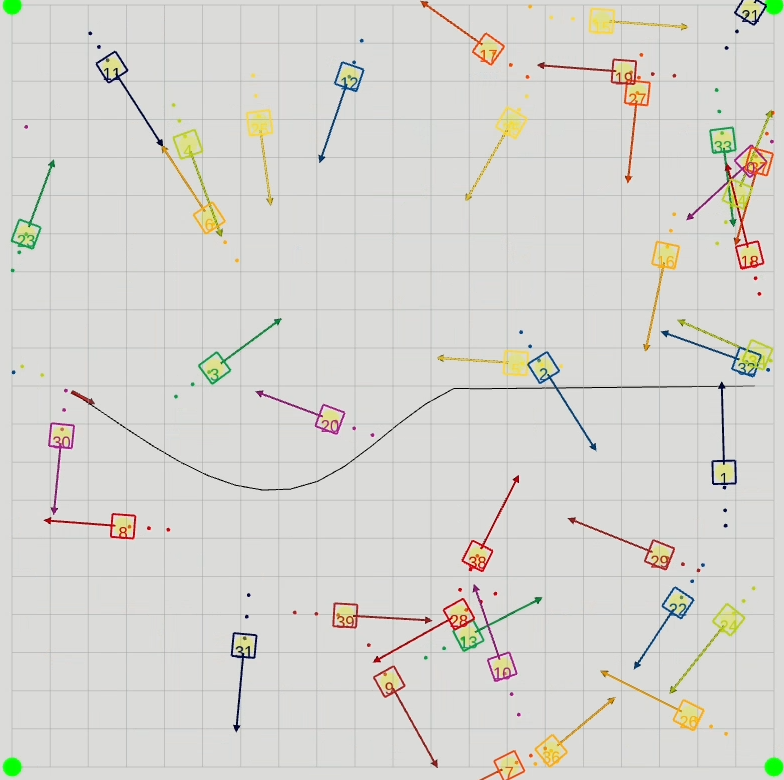}
	} 
	\subfigure[Benchmark datasets \cite{eth-dataset, ucy-dataset}.]{
		\label{real}
		\includegraphics[width=0.232\textwidth]{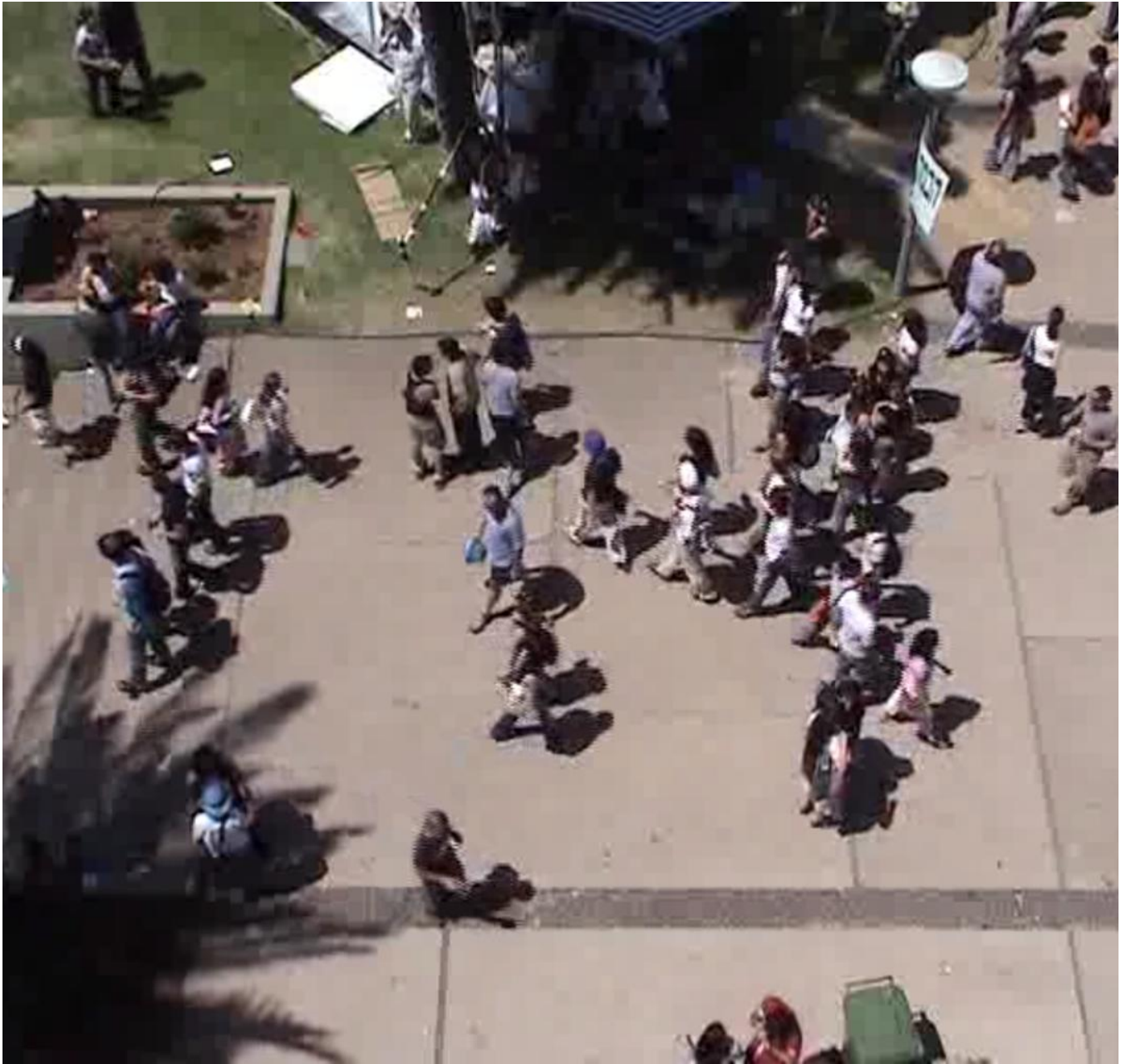}
	}
	\caption{The simulation environment and benchmark datasets.}
	\label{fig:env}
	\vspace{-4mm}
\end{figure}

The performance of motion planning with moving obstacles is mainly related to three factors: the number of obstacles, the maximum velocity of the robot, and the safe distance. To quantify their influence, we design a 10m-by-10m simulation environment, shown in Fig. \ref{sim}, to conduct control experiments. Once the obstacle moves outside the environment, it is transported to the antipodal position to maintain a constant obstacle density. The transportation of obstacles is unknown to the robot, which means the robot may meet obstacles that suddenly occur or disappear at boundaries. The moving directions of obstacles are randomly initialized, and the speeds are uniformly sampled from 1.2$\sim$1.8 m/s.

\begin{figure}[t]
	\centering
	\subfigure{
		\includegraphics[width=0.23\textwidth]{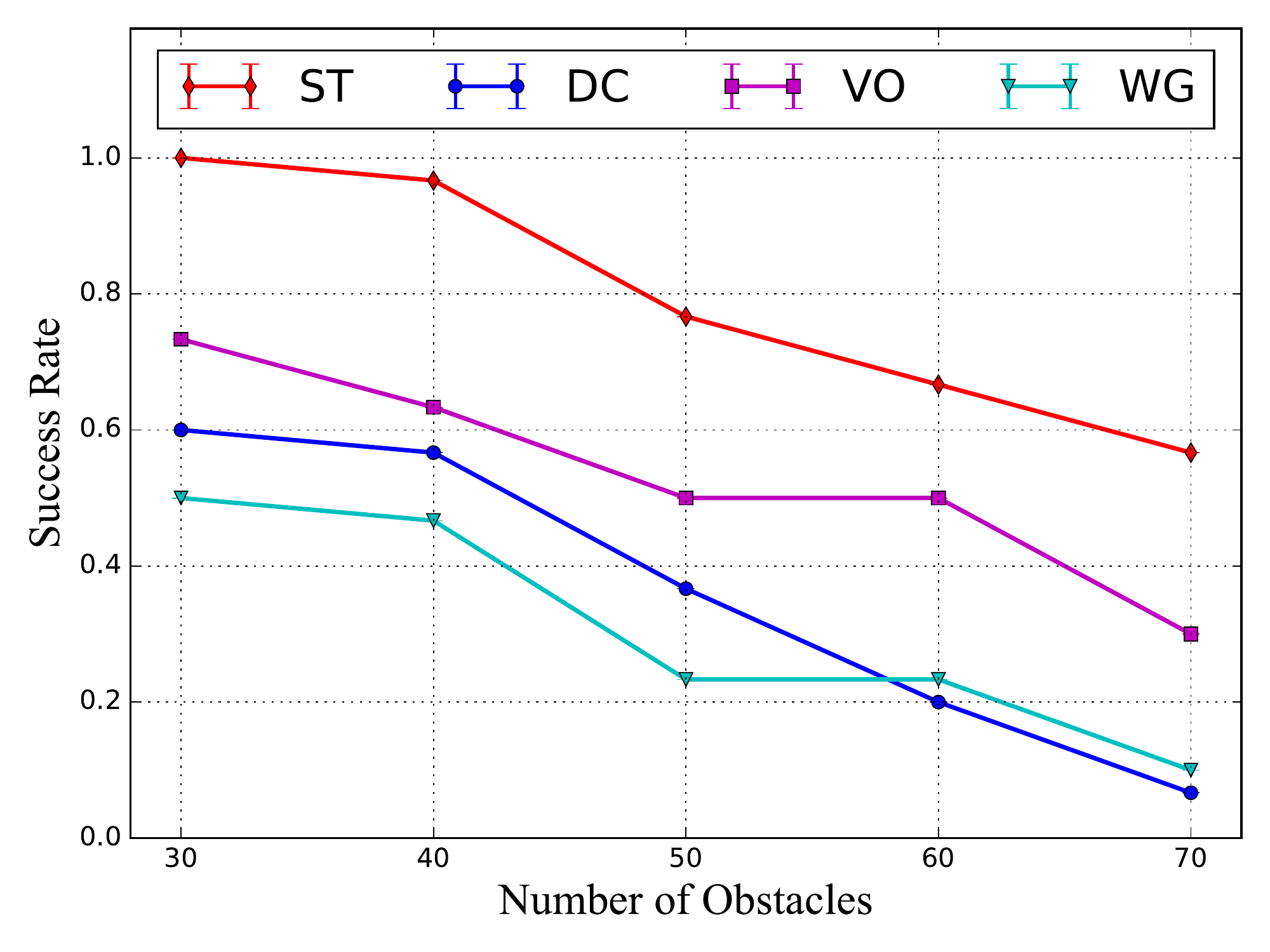}
		\includegraphics[width=0.23\textwidth]{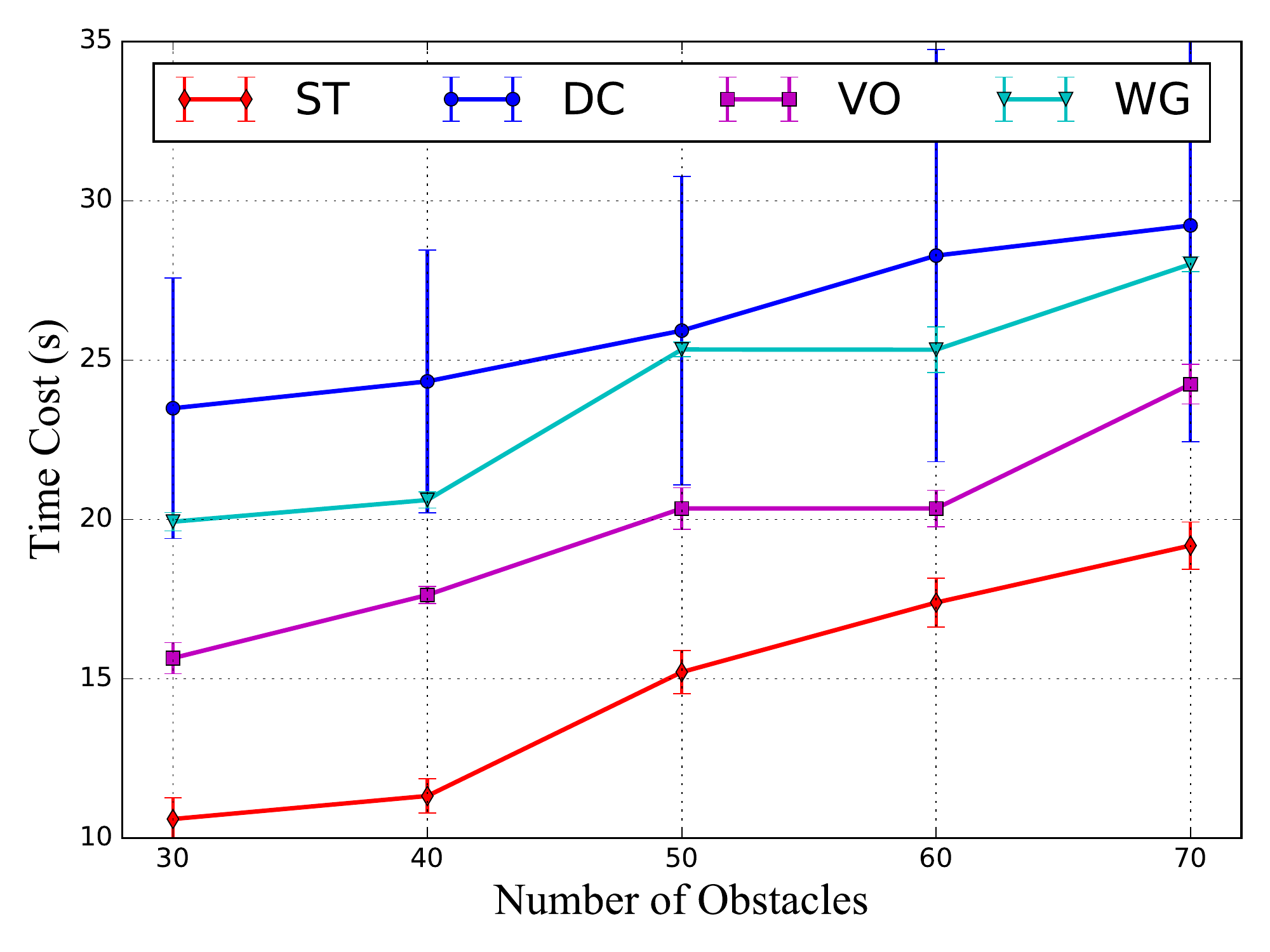}
		\label{simexp1}
	} 
	\subfigure{
		\includegraphics[width=0.23\textwidth]{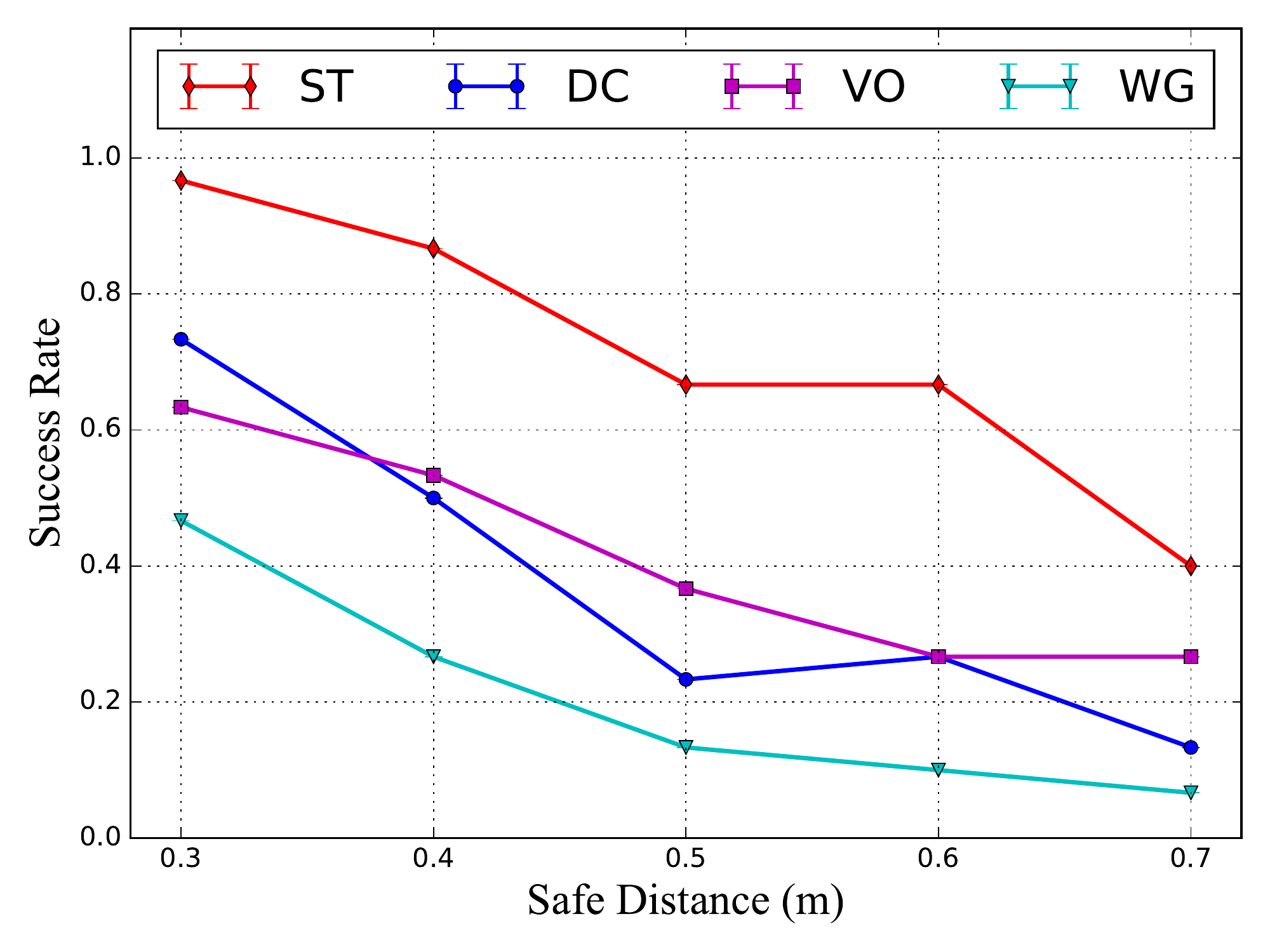}
		\includegraphics[width=0.23\textwidth]{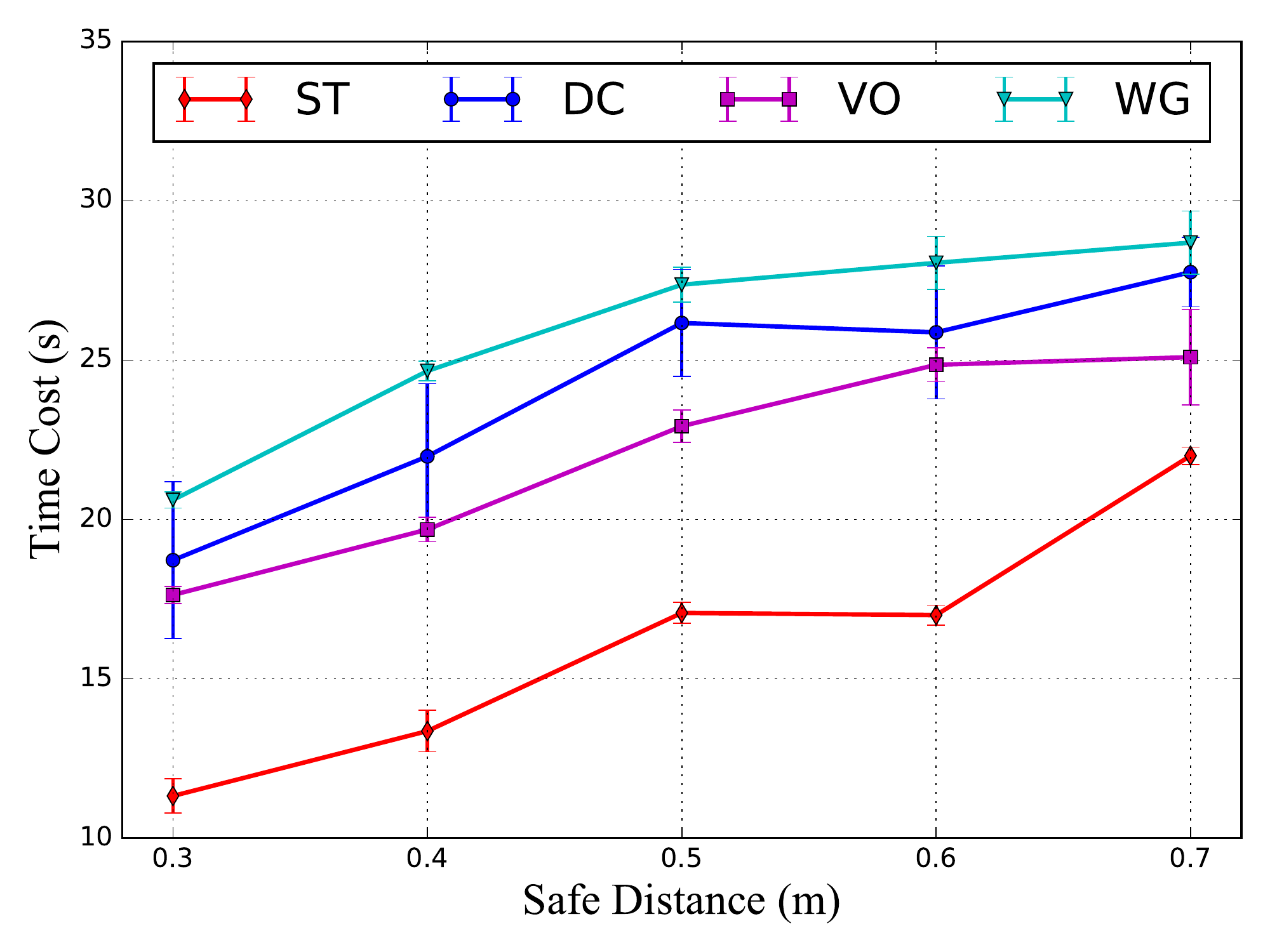}
		\label{simexp2}
	} 
	\subfigure{
		\includegraphics[width=0.23\textwidth]{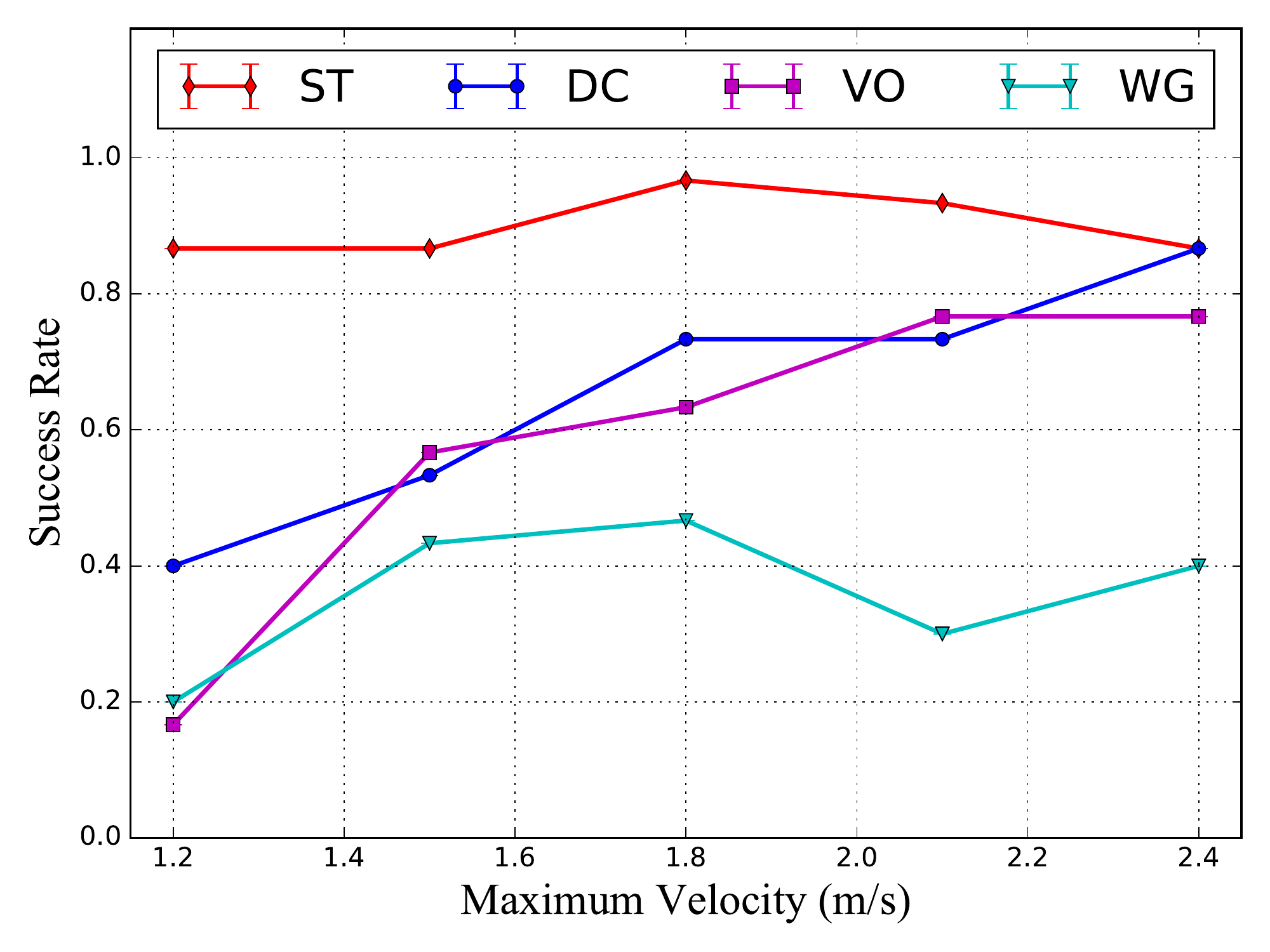}
		\includegraphics[width=0.23\textwidth]{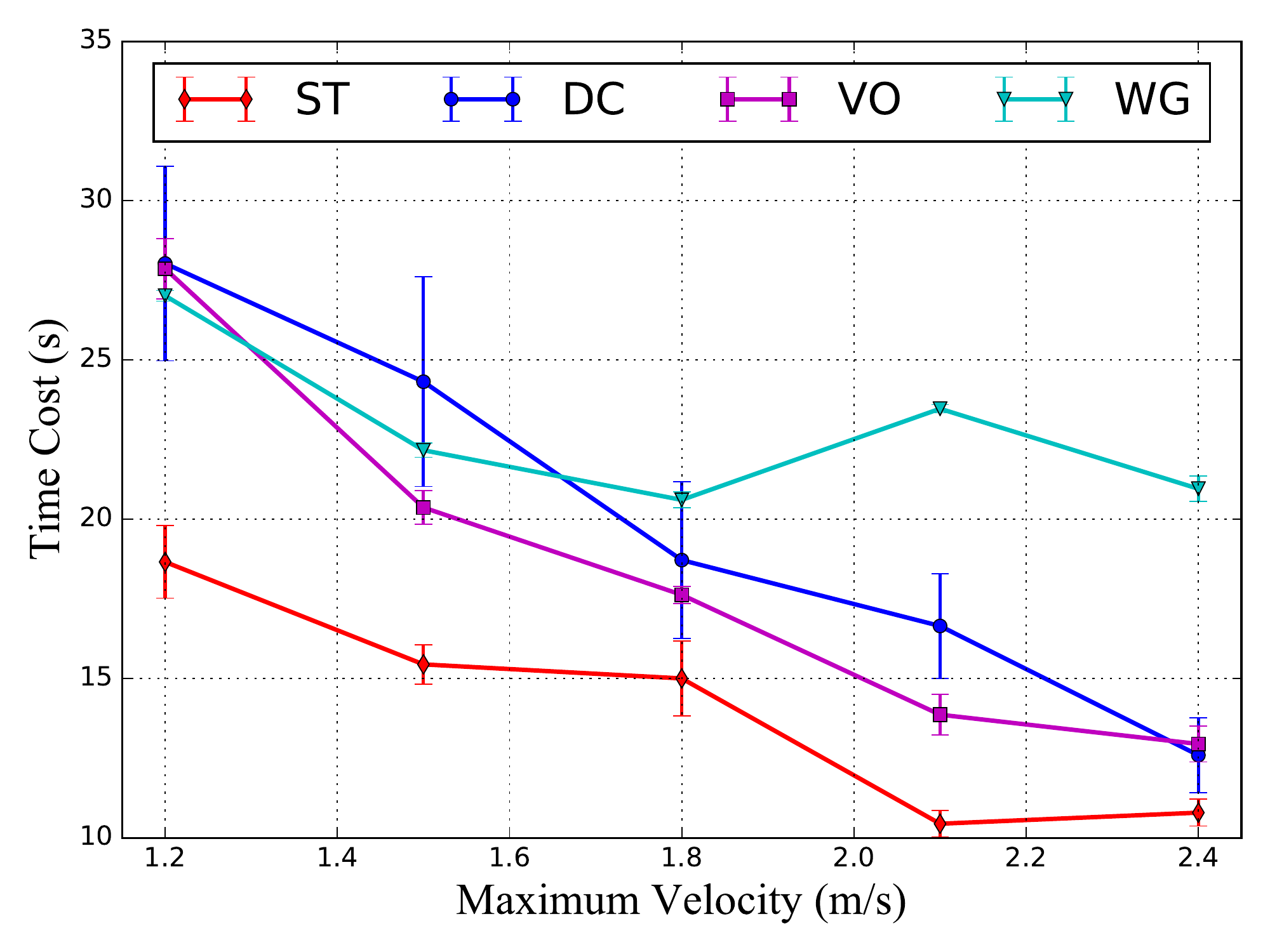}
		\label{simexp3}
	}
	\caption{Performance on simulation environment.}
	\label{fig:simexp}
	\vspace{-4mm}
\end{figure}

The default parameter settings for the experiments are as follows: 40 obstacles, 1.8m/s maximum speed in $x$ and $y$ direction, and 0.3m safe distance. In each experiment, one of the parameters is changed. The planning start and goal are set to the middle points on the left and right, respectively. Each planning starts at a random time point and repeats 30 times. The success rate and time cost are recorded as evaluation metrics. If the planner cannot reach the goal in 30 seconds, this test is counted as a failure, and the time cost is recorded as 30 seconds. If the planner fails before timeout, the time cost is also recorded as 30 seconds. The experiment results are shown in Fig. \ref{fig:simexp}.

%

\textbf{Number of obstacles}: As shown in the first row of Fig. \ref{fig:simexp}, all the planners exhibit a worse performance when the number of obstacles increases. The reason is that more admissible regions in state-time space are occupied by obstacles, making it more difficult for the planners to find a feasible solution. Compared with the baselines, the ST method maintains a higher success rate and lower time cost, which significantly demonstrates the advantage of the proposed framework. 
	
\textbf{Safe distance}: As shown in the second row of Fig. \ref{fig:simexp}, the performance changes exhibit a similar trend with those of the first experiment, which is also caused by a similar reason. Larger safe distances reduce the volume of admissible state-time space and thus increase the difficulties of planning. In this experiment, the proposed framework again demonstrates superior performance over the baselines.   
	
\textbf{Maximum velocity}: Considering the motion capabilities of the robot and safety issues, we limit the maximum speed of the robot in $x$ and $y$ directions. The experiment results are shown in the third row of Fig. \ref{fig:simexp}. As is illustrated, the success rate of DC is positively related to the maximum velocity, while the other three planners show a decreasing tendency after a certain test point. The difference is essentially caused by different implementations of the algorithms. For DC method, the collision checking is performed in a continuous space by solving a quadratic function, and thus is not sensitive to velocity changes. For ST and VO, collision checking is based on time discretization. Larger maximum velocities will decrease the checking resolution and cause unexpected collisions. Nevertheless, the ST method still exhibits a better performance over the baselines.

\subsection{Experiments on Benchmark Dataset}
\begin{figure}[t]
	\centering
	\subfigure{
		\includegraphics[width=0.36\textwidth]{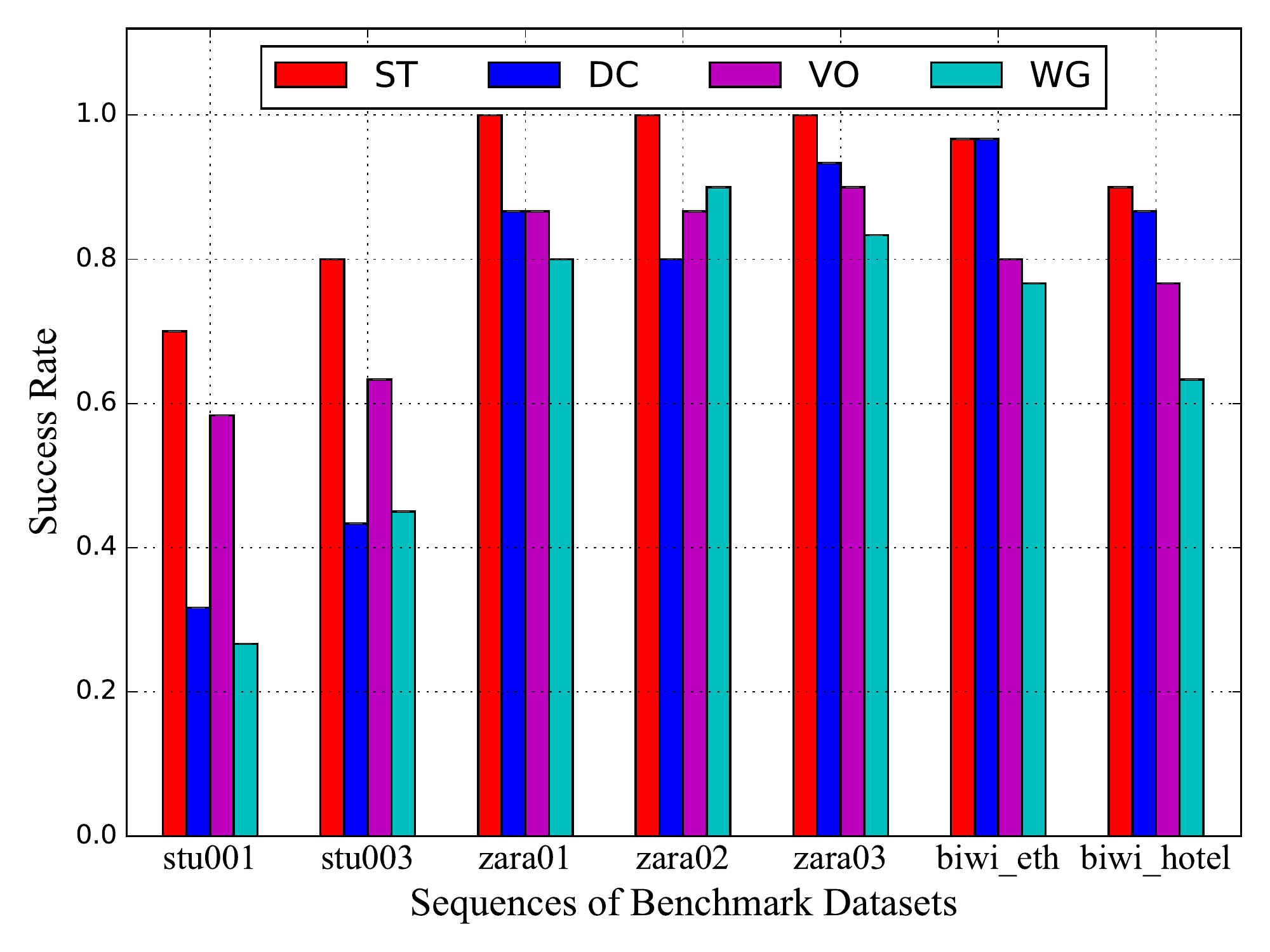}
	} 
	
	\subfigure{
		\includegraphics[width=0.36\textwidth]{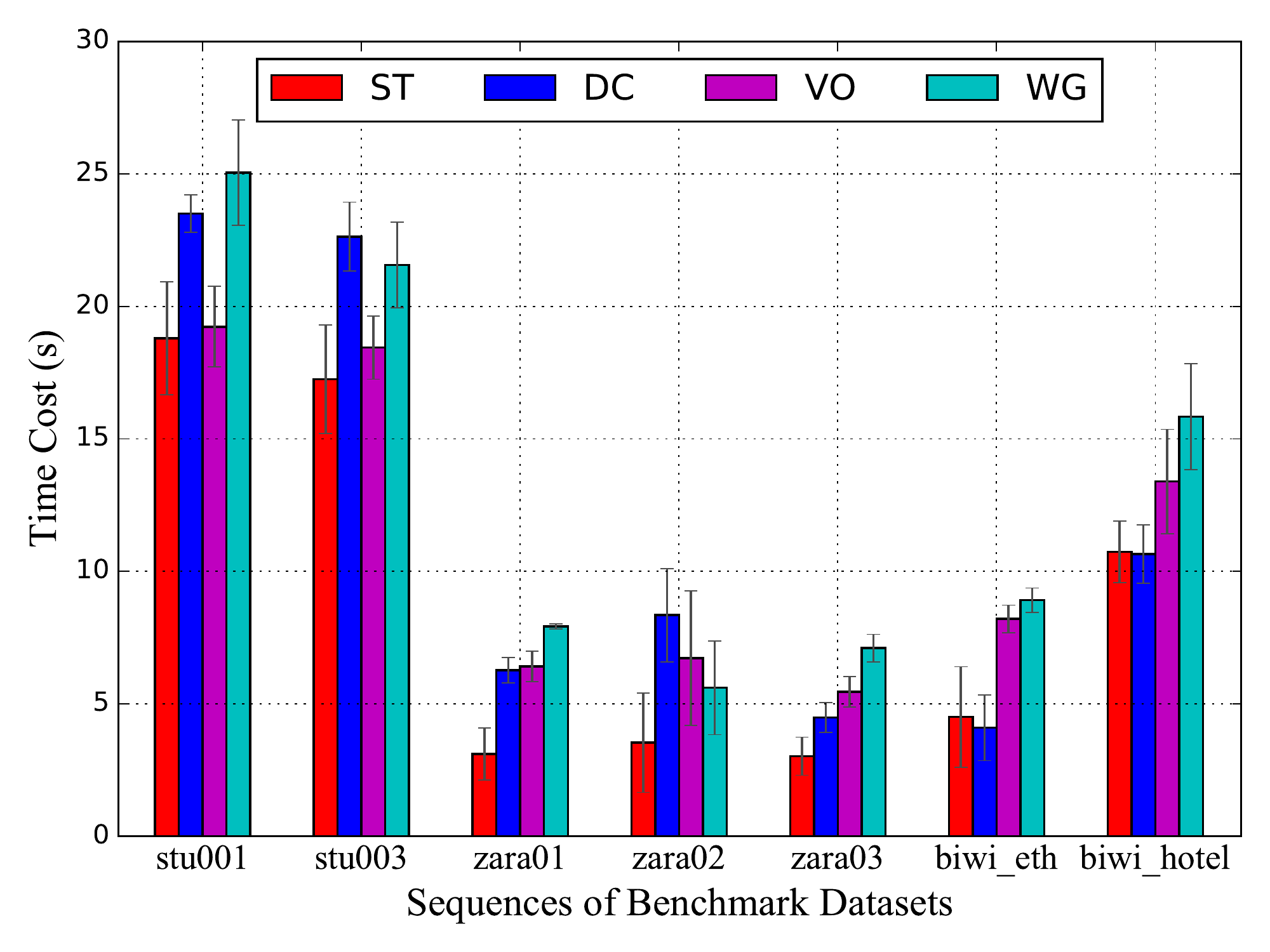}
	}
	\caption{Performance on benchmark datasets.}
	\label{fig:realexp}
		\vspace{-4mm}
\end{figure}

\begin{figure*}[t]
	\centering
	\subfigure[stu001]{
		\includegraphics[width=0.193\textwidth]{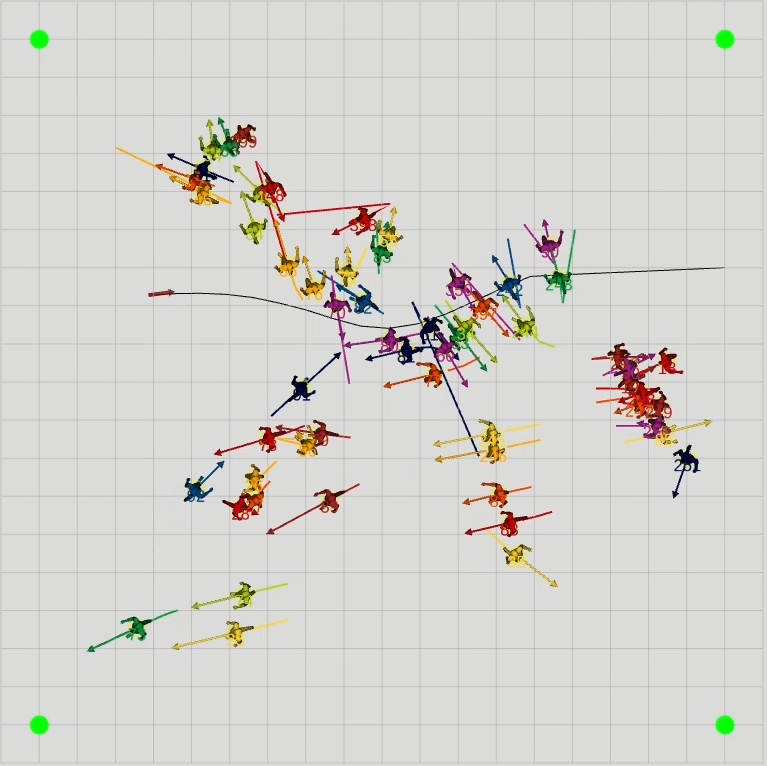}
	}\hspace{-3mm}
	\subfigure[stu003]{
		\includegraphics[width=0.194\textwidth]{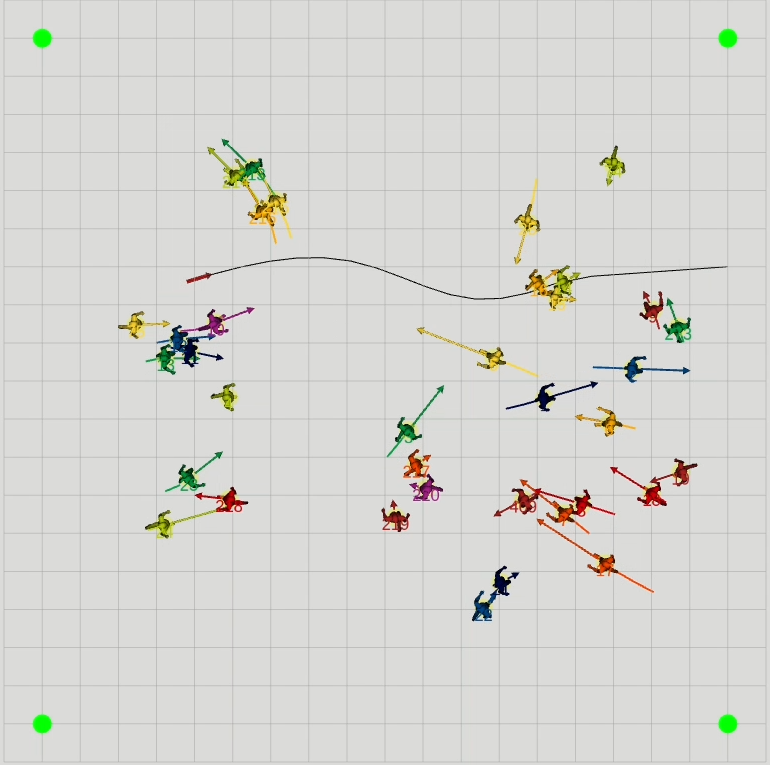}
	}\hspace{-3mm}
	\subfigure[zara01]{
		\includegraphics[width=0.192\textwidth]{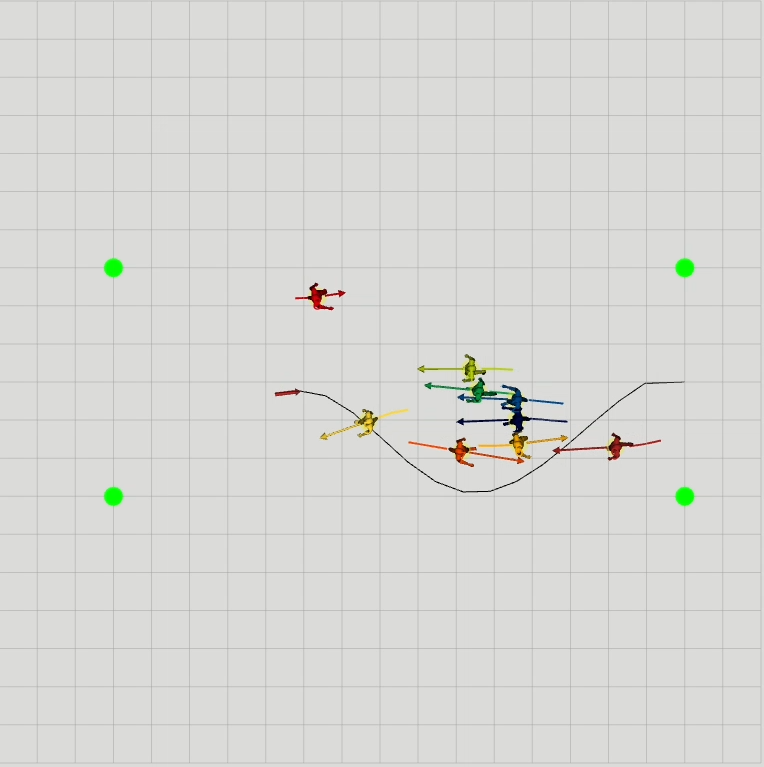}
	}\hspace{-3mm}
	\subfigure[zara03]{
		\includegraphics[width=0.1936\textwidth]{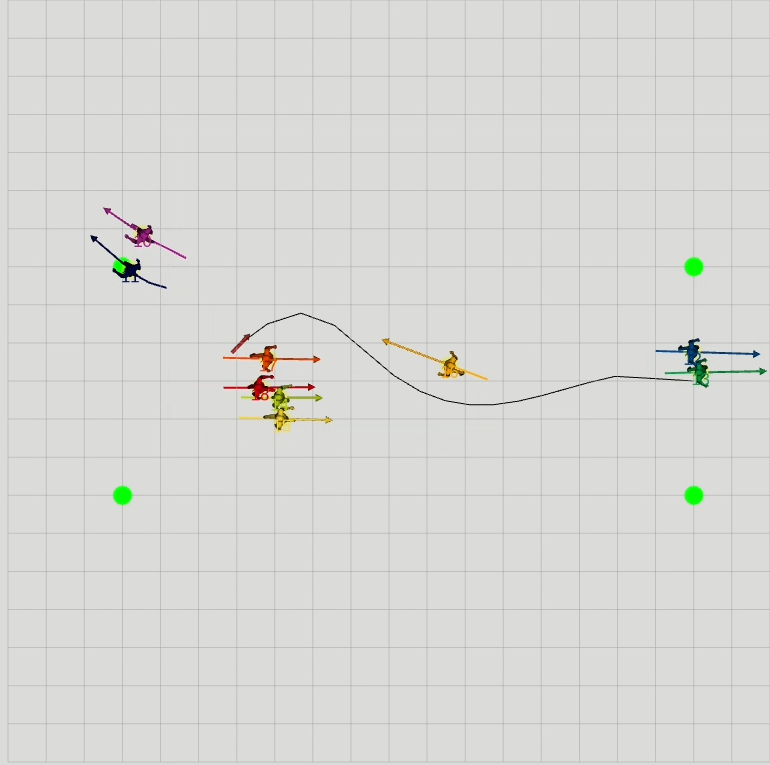}
	}\hspace{-3mm}
	\subfigure[biwi\_eth]{
		\includegraphics[width=0.1931\textwidth]{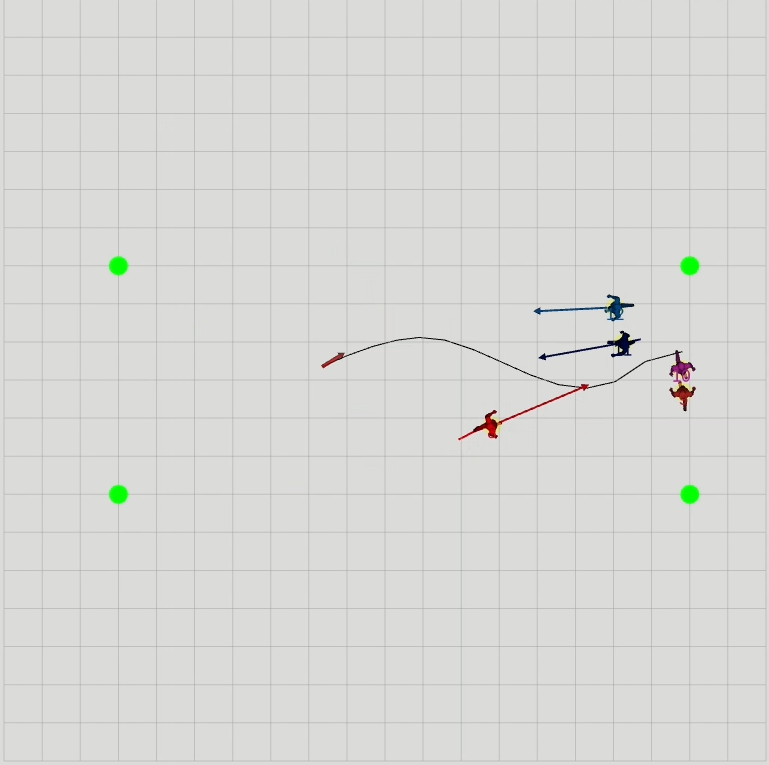}
	}   	
	\caption{The planning results of our online state-time planner on different data sequences.  }
	\label{fig:demo}
	\vspace{-4mm}
\end{figure*}

For benchmarking the proposed framework, three publicly available datasets from \cite{eth-dataset, ucy-dataset} are adopted. There are seven sequences of pedestrian videos captured in different scenarios, denoted by \textit{stu001, stu003, zara01, zara02, zara03, biwi\_eth}, and \textit{biwi\_hotel}. All the sequences are interpolated and then played at 10Hz. The first two sequences are captured from densely populated environments, and the last two only contain very sparse pedestrians, as shown in Fig. \ref{fig:demo}. To make sure the planning is valid, we add a boundary for each sequence, denoted by the green circles in Fig. \ref{fig:demo}. The start and goal are set to the middle left and middle right, respectively, and the maximum velocity is set to 1.5m/s in both x and y directions. The safe distance is set to 0.4m. For each sequence, 30 tests are performed, and each test starts at a random time point of the dataset. The same evaluation metrics with the simulation experiment are adopted, and the results are shown in Fig. \ref{fig:realexp}. 

It can be seen that the ST and VO methods have higher success rates than the DC and WG methods on \textit{stu001} and \textit{stu003} sequences, while the ST and DC methods perform better on \textit{biwi\_eth} and \textit{biwi\_hotel} sequences. The ST method succeeds in all the tests on \textit{zara} sequences, while the DC and VO methods show competitive performance. 
The performance variance is essentially caused by the different properties of the datasets, and certainly, the different planner implementations. The ST and VO methods take into account obstacle motions for velocity planning, and thus can tackle more complected cases. The DC method performs collision checking based on a static topology of the obstacles during velocity extrapolation, which however varies much in highly dynamic environments. This makes the DC method not suitable for handling complicated cases like the \textit{stu001} and \textit{stu003} sequences. On the other hand, the DC methods have a larger planning horizon than the VO and WG methods, thus exhibiting better time optimality and higher success rate in sparse sequences.       

Compared with baseline methods, the proposed ST method conducts state-time planning with a long horizon, and combines the long-term and short-term information of the environments, thus exhibiting significant advantages over the other methods. Some planning results by the ST method are shown in Fig. \ref{fig:demo}, which demonstrates its capability of generating smooth and safe trajectories. The proposed ST method is also very efficient. On the most challenging \textit{stu001} and \textit{stu003} sequences, the front end and back end are with an average time cost of 1.6ms and 15ms, respectivey.    


\section{Conclusions}
In this work, we develop an online state-time trajectory planner that is suitable for planning in highly dynamic environments. Experiments on the simulation environment and benchmark datasets significantly demonstrate the advantages of our proposed planner. The integration of path initialization by the front end and the Timed-ESDF in the back end make a great contribution to the high success rate of the state-time planning. The current planner does not consider holonomic constraints, which will be the focus of our future work.

\bibliographystyle{IEEEtran}
\bibliography{ref}

%


%
%




\end{document}